%% file: bmvc_final.tex
\documentclass{bmvc2k}

\usepackage{times}
\usepackage{epsfig}
\usepackage{graphicx}
\usepackage{amsmath}
\usepackage{amssymb}
\usepackage{bm}
\usepackage{multirow}
\usepackage{bbold}
\usepackage[percent]{overpic}

\input{inc/defs}

\title{\papertitle}
\credit{}

\runninghead{Sulc A.~\etaltit{}}{\papertitle}

\begin{document}

\maketitle

\begin{abstract}
  \input{tex/abstract}
\end{abstract}

\vspace{-0.4cm}
\section{Introduction}
\vspace{-0.2cm}
\label{sect:introduction}
\input{tex/introduction}

\section{Related Work}
\vspace{-0.2cm}
\label{sect:rw}
\input{tex/rw}
\section{Refraction Through a Surface}
\vspace{-0.2cm}
\label{sect:bg} 
\input{tex/background}

\section{Optimization}
\vspace{-0.2cm}
\label{sect:optim}
\input{tex/optim}

\section{Experiments}
\vspace{-0.2cm}
\label{sect:experiments}
\input{tex/experiments}

\section{Conclusions}
\vspace{-0.2cm}
\label{sect:conclusion}
\input{tex/conclusion}
\clearpage

\section*{Acknowledgement}
\label{sect:ack}
\input{tex/ack}
\bibliography{egbib}

\end{document}

\maketitle

\section{More Detailed Evaluations}
\begin{figure}
    \centering
    \resizebox{1.0\linewidth}{!}{
    \begin{tabular}{ccc}
        & \multicolumn{2}{c}{\benchmarkname{wave1}}\\
        & \RMSE{} & \MAE{} \\
        \rotatebox{90}{\hspace{3em}Indep. init.} &
        \includegraphics[width=0.5\linewidth]{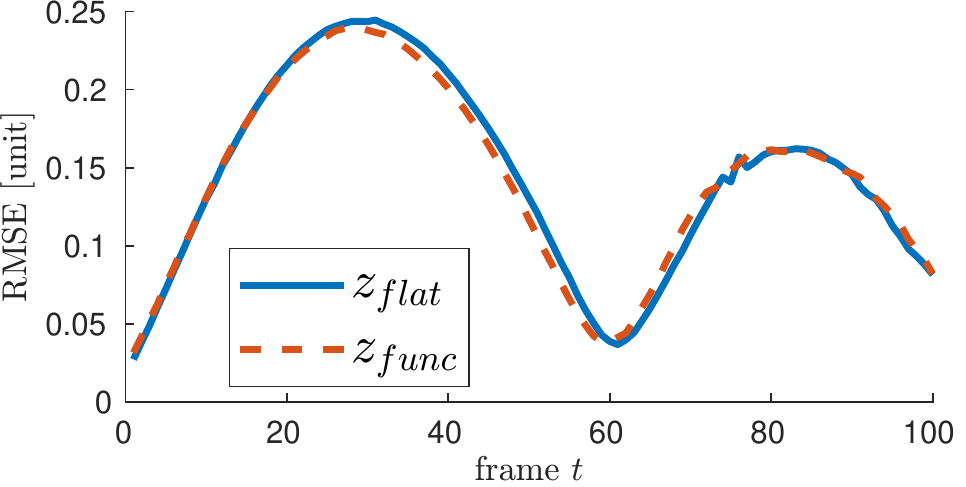} & \includegraphics[width=0.5\linewidth]{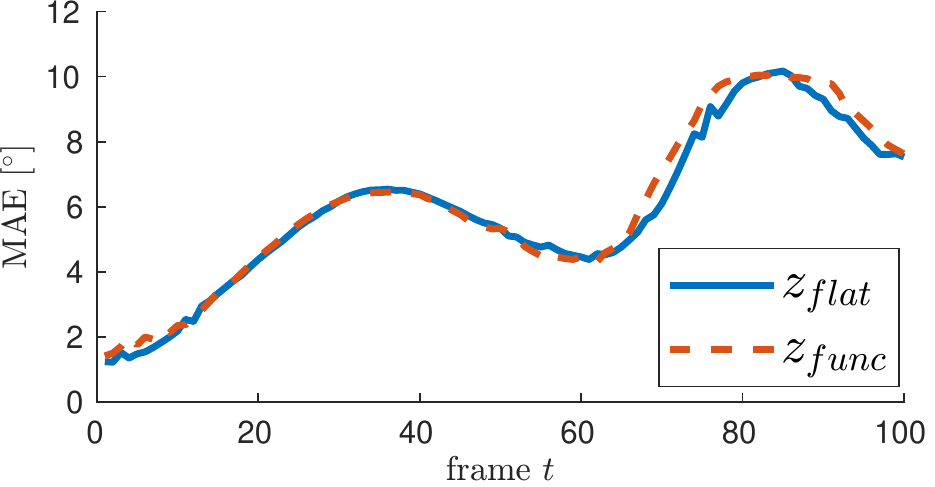}\\
        \rotatebox{90}{\hspace{3em}Seq. init.} & 
        \includegraphics[width=0.5\linewidth]{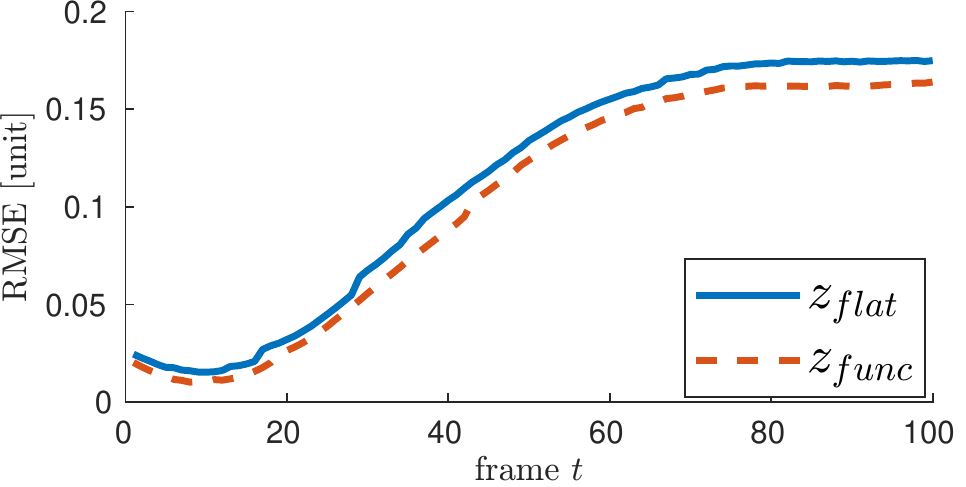} & \includegraphics[width=0.5\linewidth]{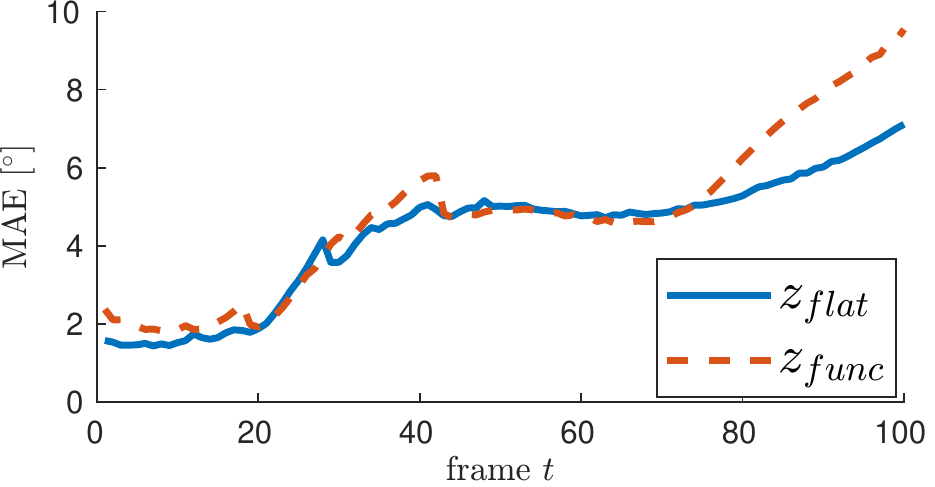}\\\
        \rotatebox{90}{\hspace{3em}Init. at $\dev{0} = 2.0$} &
        \includegraphics[width=0.5\linewidth]{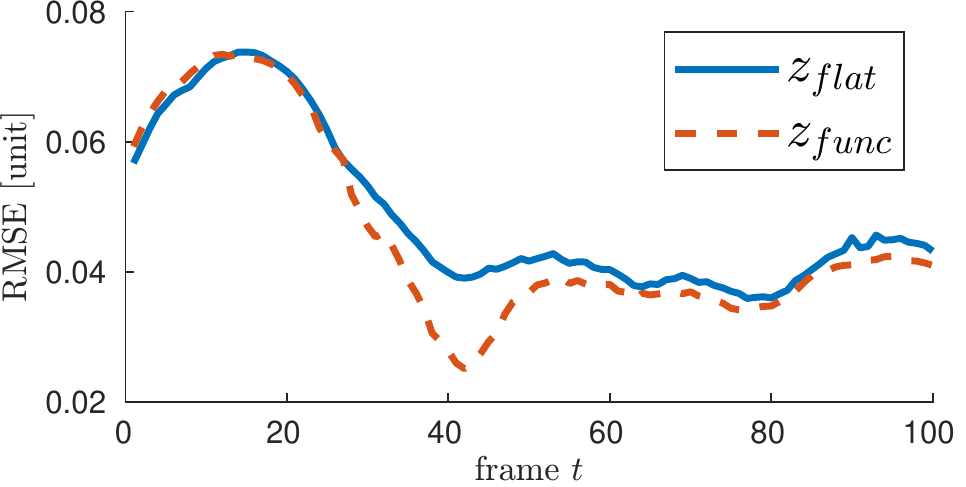} & \includegraphics[width=0.5\linewidth]{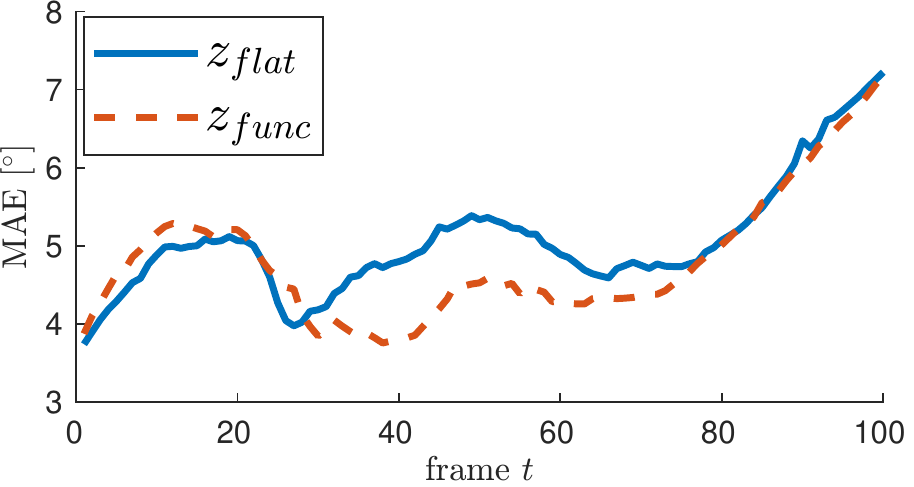}
    \end{tabular}}
    \caption{Frame-wise~\RMSE{} and~\MAE{} evaluation of~\benchmarkname{wave1} with different initialization schemes.
    The left column shows \RMSE{} of estimated depth $\hat{z}$ from the ground truth depth.
    The right column shows \MSE{} of estimated normals $\n{}$ from the ground truth.
    The first row shows results where each frame is initialized with our scheme independently.
    The second row shows results where each frame is initialized with the estimate from the previous frame except the first, which is initialized with our scheme.
    The third row shows results where each frame is initialized with an identical $\dev{} = 2$ which is close to the mean location of the ground truth surface.
    }
    \label{fig:wave1}
\end{figure}

\begin{figure}
    \centering
    \resizebox{1.0\linewidth}{!}{
    \begin{tabular}{ccc}
        & \multicolumn{2}{c}{\benchmarkname{wave2}}\\
        & \RMSE{} & \MAE{} \\
        \rotatebox{90}{\hspace{3em}Indep. init.} &
        \includegraphics[width=0.5\linewidth]{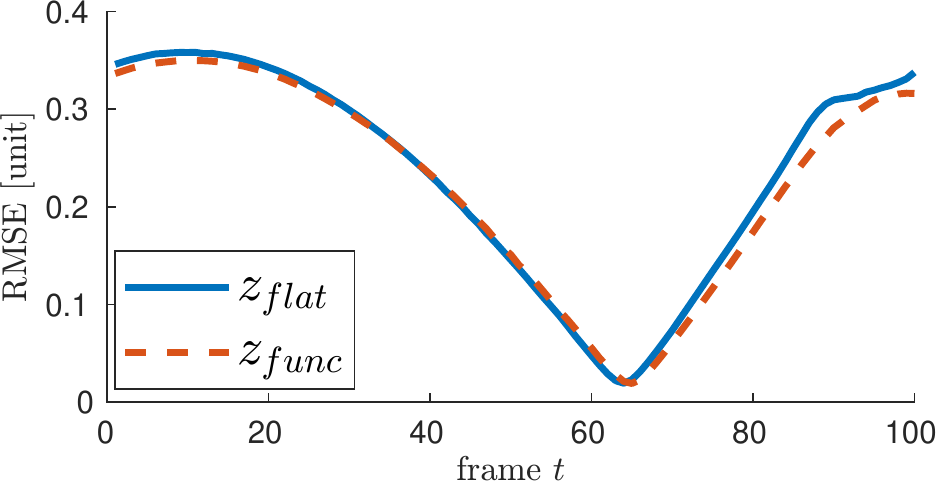} & \includegraphics[width=0.5\linewidth]{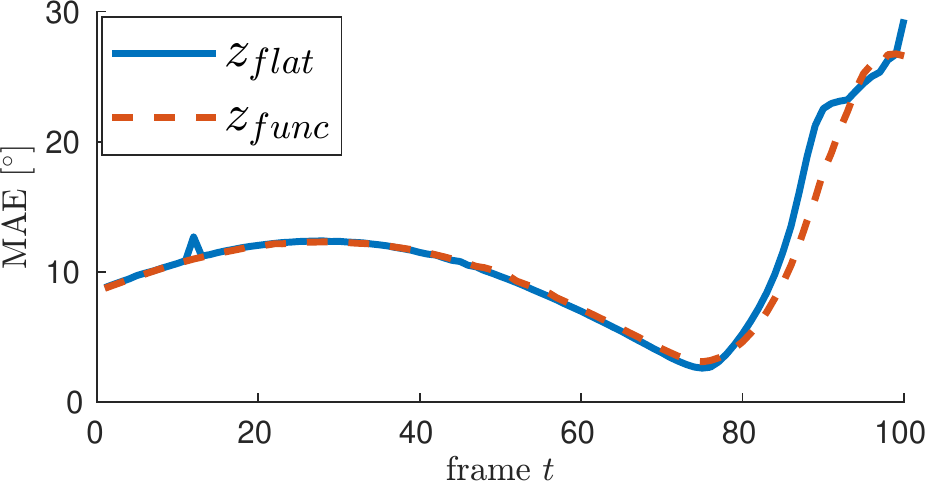}\\
        \rotatebox{90}{\hspace{3em}Seq. init.} & 
        \includegraphics[width=0.5\linewidth]{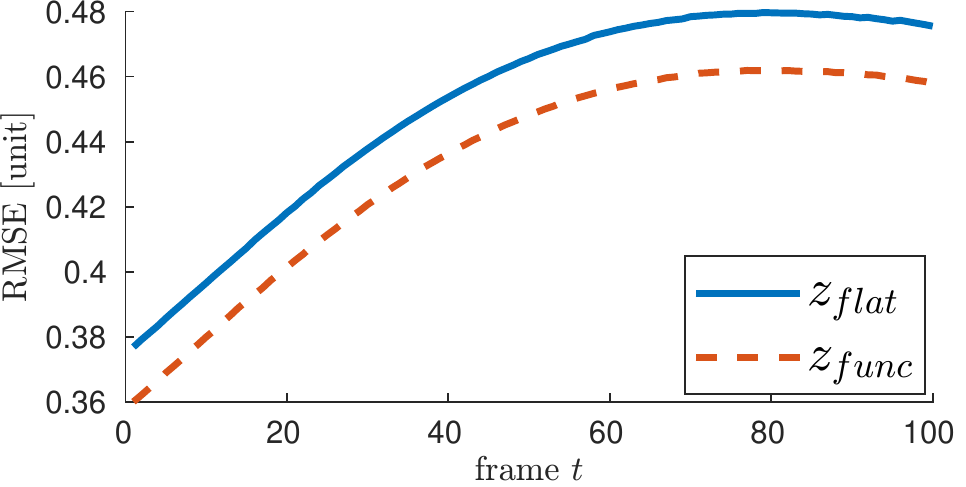} & \includegraphics[width=0.5\linewidth]{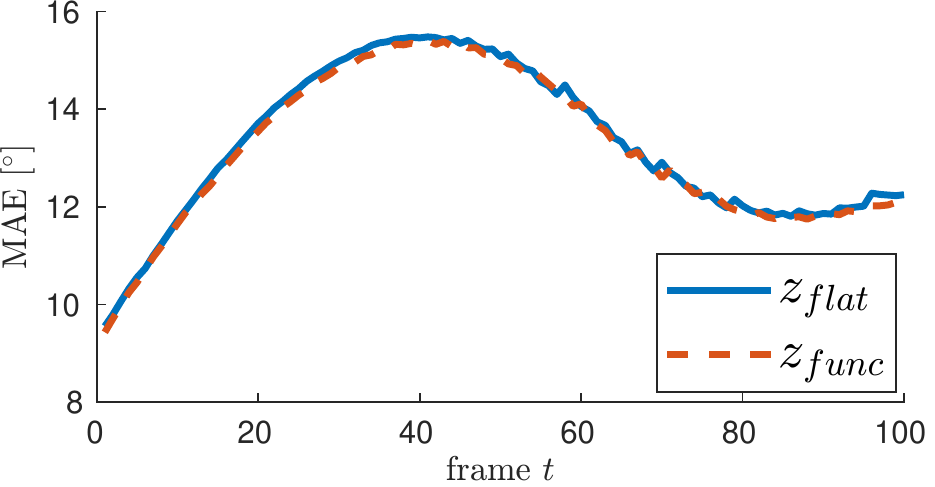}\\\
        \rotatebox{90}{\hspace{3em}Init. at $\dev{0} = 2.0$.} &
        \includegraphics[width=0.5\linewidth]{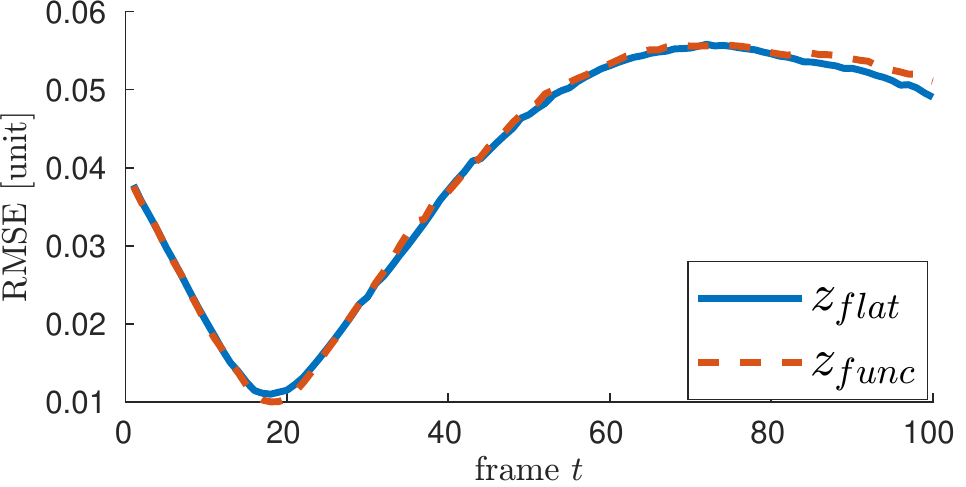} & \includegraphics[width=0.5\linewidth]{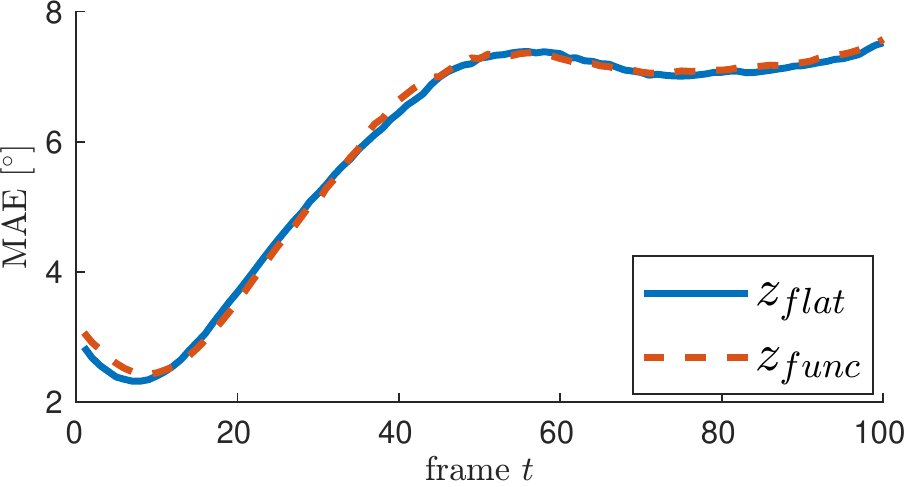}
    \end{tabular}}
    \caption{Frame-wise~\RMSE{} and~\MAE{} evaluation of~\benchmarkname{wave2} with different initialization schemes.
    The left column shows \RMSE{} of estimated depth $\mathbf{z}$ from the ground truth depth.
    The right column shows \MSE{} of estimated normals $\n{}$ from the ground truth.
    The first row shows results where each frame is initialized with our scheme independently.
    The second row shows results where each frame is initialized with the estimate from the previous frame except the first, which is initialized with our scheme.
    The third row shows results where each frame is initialized with an identical $\dev{} = 2$ which is close to the mean location of the ground truth surface.
    }
    \label{fig:wave2}
\end{figure}

\fig{fig:wave1} and \fig{fig:wave2} show detailed frame-wise~\RMSE{} and~\MAE{} results on \benchmarkname{wave1} and \benchmarkname{wave2} respectively. 

\section{Visualizations}

Images from and results of our real-world experiment are in the attached video file \fname{rw.avi}.
In each frame, the left image shows the background image in the first frame and input images in all further frames. The right image shows the reconstructed surface.

% \clearpage
% \bibliography{egbib}

%% file: inc/defs.tex
\newcommand{\X}[1]{X_{#1}}
\newcommand{\Xb}[1]{B_{#1}}
\newcommand{\Xr}[1]{\X{#1}}
\newcommand{\LOS}[1]{\boldsymbol{l}_{#1}}
\newcommand{\LOSr}{\LOS{}}
\newcommand{\n}[1]{\boldsymbol{n}_{#1}}
\newcommand{\s}[1]{\boldsymbol{s}_{#1}}
\newcommand{\x}[1]{\boldsymbol{x}_{#1}}
\newcommand{\OF}[1]{\boldsymbol{u}_{#1}}
\newcommand{\xb}{\x{B}}
\newcommand{\xr}{\x{}}
\newcommand{\calb}{K}
\newcommand{\de}[1]{d_{#1}}
\newcommand{\dev}[1]{\boldsymbol{d}_{\text{#1}}}
\newcommand{\devgt}[1]{\dev{\text{gt}}}
\newcommand{\devh}[1]{\dev{\text{h}}}

\newcommand{\snell}[3]{\text{snell}\left(#1,#2,#3\right)}
\newcommand{\len}[1]{E_{#1}}
\newcommand{\eq}[1]{Eq.~\eqref{#1}}
\newcommand{\tab}[1]{Tab.~\ref{#1}}
\newcommand{\fig}[1]{Fig.~\ref{#1}}

\newcommand{\sect}[1]{Sec.~\ref{#1}}
\usepackage{xcolor}

\newcommand{\benchmarkname}[1]{\emph{#1}}
\newcommand{\MAE}{\text{MAE}}
\newcommand{\RMSE}{\text{RMSE}}

\newcommand{\zflat}{z_{\text{flat}}}
\newcommand{\zfunc}{z_{\text{func}}}
\newcommand{\zone}{z_{\text{1}}}
\newcommand{\ztwo}{z_{\text{2}}}

\newcommand{\ie}{i.e.}

\newcommand{\unit}{unit}
\newcommand{\new}[1]{{#1}}

\def\etal{\emph{et al}\bmvaOneDot}
\def\etaltit{{et al}\bmvaOneDot}

\newcommand{\papertitle}{Towards Monocular Shape from Refraction}
\newcommand{\credit}{
\addauthor{Antonin Sulc}{antonin.sulc@uni-konstanz.de}{1}
\addauthor{Imari Sato}{imarik@nii.ac.jp}{2}
\addauthor{Bastian Goldluecke}{bastian.goldluecke@uni-konstanz.de}{1}
\addauthor{Tali Treibitz}{ ttreibitz@univ.haifa.ac.il}{3}

\addinstitution{
 Computer Vision and Image Analysis\\
 University of Konstanz\\
 Konstanz, Germany
}
\addinstitution{
 National Institute of Informatics\\
 Tokyo, Japan
}
\addinstitution{
Charney School of Marine Sciences,\\
 University of Haifa,\\
 Haifa, Israel
}}

\newcommand{\makewaveimagecell}[2]{\begin{overpic}[width=0.2\linewidth]{local_images/experiments/#1} \put (0,70) {\large{#2}} \end{overpic}}

%% file: tex/abstract.tex
Refraction is a common physical phenomenon and has long been researched in computer vision. Objects imaged through a refractive object appear distorted in the image as a function of the shape of the interface between the media. This hinders many computer vision applications, but can be utilized for obtaining the geometry of the refractive interface. Previous approaches for refractive surface recovery largely relied on various priors or additional information like multiple images of the analyzed surface. In contrast, we claim that a simple energy function based on Snell's law enables the reconstruction of an arbitrary refractive surface geometry using just a single image and known background texture and geometry. In the case of a single point, Snell's law has two degrees of freedom, therefore to estimate a surface depth, we need additional information. We show that solving for an entire surface at once introduces implicit parameter-free spatial regularization and yields convincing results when an intelligent initial guess is provided. We demonstrate our approach through simulations and real-world experiments, where the reconstruction shows encouraging results in the single-frame monocular setting.

%% file: tex/introduction.tex
When we look at objects through a refractive surface like water or glass, the object changes appearance, and its image is distorted (see~\fig{fig:teaser}).
This distortion is caused by refraction that happens when light propagates to a medium that transmits light at a different speed.
When the ray's path is obstructed with a refractive material, it causes a change of the light ray's trajectory on the interface based on its surface geometry.
%
% This non-linear change of the trajectory is described by Snell's law~\cite{kidger2001fundamental}.

\input{local_figures/teaser}

This situation happens, for example, when looking into a body of water from above with a drone, or as a lifeguard at a pool. Reconstructing the surface can help compensate for its effects to see through it, and several applications can exploit the surface shape itself, such as sea condition monitoring or refractive object inspection.

While refraction can confuse many computer vision algorithms, the bending of the light also provides hints towards the shape of the surface. In the problem of reconstruction of such surfaces, each surface point has two unknowns: its location and its normal. Previously it was mostly solved in settings with multiple views~\cite{qian2018simultaneous,qian20163d,kutulakos2008theory} or multiple frames~\cite{thapa2020dynamic}. We consider a simpler setup with a single frame monocular camera with a known arbitrary background.

We ask, given an object point and its refracted image, can we determine the light path between them without knowing anything about the surface or its location? While for a single point this problem is ill-posed, we show that solving this problem for all points at once imposes implicit spatial regularization and yields reconstruction of the full 3D of the refractive surface.

We propose an energy formulation such that we seek a surface that explains the refracted projection of the background object using Snell's law. 
To overcome the ambiguity, we optimize the whole surface in a single energy, which allows us a calculation of the necessary normal field.
Furthermore, our energy works directly in world coordinates and minimizes the geometric errors such that we can obtain a full 3D reconstruction of the sought surface. 

\vspace{0.2cm}
\noindent\textbf{Our contributions are:}
\vspace{-0.2cm}
\begin{itemize}
\setlength\itemsep{0pt}
\setlength\itemindent{0pt}
\item[--] We present a method that can estimate a complete 3D geometry of a refractive surface from a single distorted image with knowledge of its background texture and arbitrary depth using a standard off-the-shelf camera.
\item[--] Our method relies only on the physics of refraction, and the energy formulation is parameter-free.
\end{itemize}

The rest of the paper is organized as follows: In~\sect{sect:rw} we summarize existing works. In~\sect{sect:bg}, we show the preliminaries necessary to understand our approach. In \sect{sect:optim} we show how we optimize the unknown refractive surface via our proposed energy and in~\sect{sect:experiments} we demonstrate several experiments and evaluations on synthetic and real-world scenes.

%% file: local_figures/teaser.tex
\begin{figure}[t]
    \centering
    % \begin{tabular}{cc}
    %     \small{Empty surface}& \small{Refractive Wavy Surface}\\
    %     \includegraphics[width=0.45\linewidth]{local_images/teaser/aside-empty.png} & 
    %     \includegraphics[width=0.45\linewidth]{local_images/teaser/aside-refr.png}
    % \end{tabular}
    % \includegraphics[width=1.0\linewidth]{local_images/teaser/aside-half-half.pdf}
    \resizebox{1.0\linewidth}{!}{
    \begin{tabular}{cc}
    \includegraphics[width=0.5\linewidth]{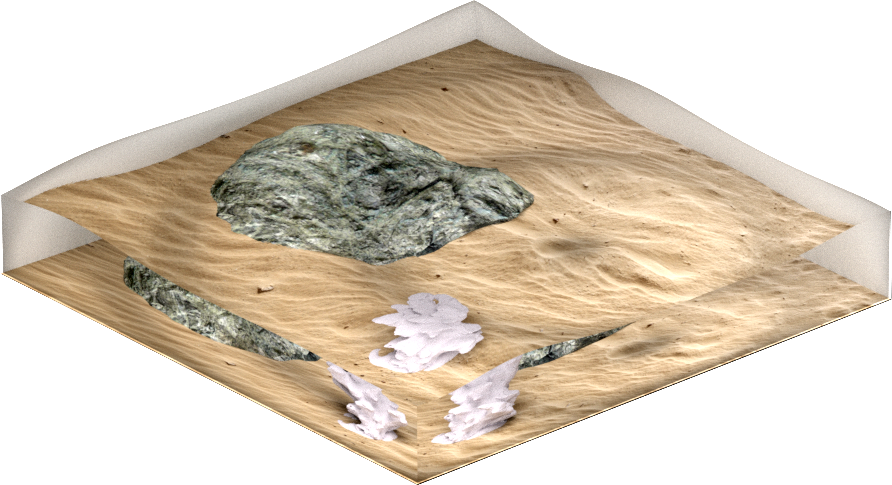} &
    \includegraphics[width=0.5\linewidth]{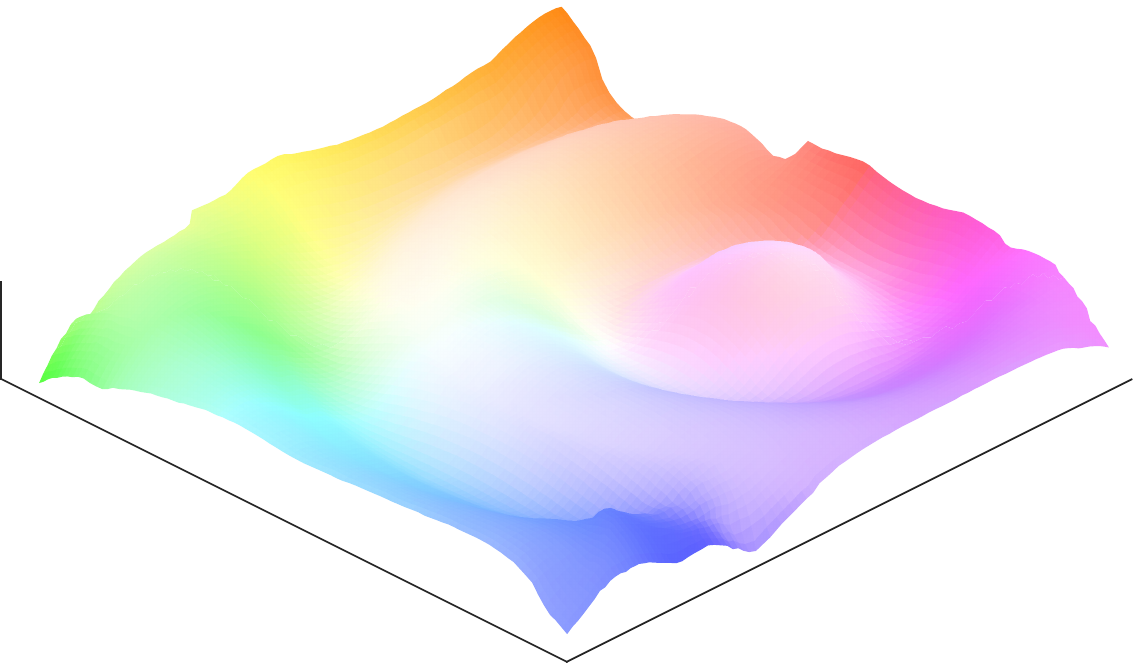}
    \end{tabular}}
    \caption{
    % Many computer vision algorithms are designed to process scenes under assumptions of \emph{homogeneous} environment between an observer and an object. 
    %
    % Under this premise, a light ray emanated from an object is transmitted to the observer in a straight path.
    % 
    % The trajectory is changed when a refractive medium is set among the object and observer and the image of the original object is deformed due to the refraction.
    % 
    % We exploit the refraction to reveal the surface depth from the image distortions.
    % 
    [Left]~A simulated scene with a refractive surface.
    We can observe a significant distortion of the background from the refraction, which is a function of the surface geometry.
    We exploit this relation and estimate the 3D geometry of the refractive surface from its image, knowing the background.
    [Right]~An example of reconstruction using our approach.}
    \label{fig:teaser}
\end{figure}

%% file: tex/rw.tex
A comprehensive analysis of the problem of refractive geometry in computer vision can be found in~\cite{ihrke2010transparent}. Here we review works related to our method.

\noindent\textbf{Multi-view Approaches.}
Ben-Ezra~\etal{}~\cite{ben2003does} presented a parametric method for shape and pose recovery of transparent objects from a known observer's motion.
The seminal work of Kutulakos~\etal{}~\cite{kutulakos2008theory} formulates conditions about the number of views and reference points under which an arbitrarily shaped specular or refractive scene can be reconstructed. They show that a surface of a refractive object can be retrieved from two viewpoints and one reference point.
Chang~\etal{}~\cite{chang2011multi} model the distortion from refraction as a depth-dependent function and reconstruct a refracted scene from multiple views.
%incorporate distortion from refraction, and by re-projecting the refracted light rays and modelling their distortion as a depth-dependent function, they reconstruct a refracted scene from multiple views.
%
Han~\etal{}~\cite{han2015fixed} showed a shape estimation method based on altering a background pattern of an object immersed in water. % and by triangulating light paths they estimated depth.
The work of Morris~\etal{}~\cite{morris2011dynamic} introduces the reconstruction of dynamic refractive surfaces with an unknown refractive index by exploiting the refractive disparity in the multi-frame stereo setting.
In~\cite{alterman2016triangulation} refraction through a dynamic surface was treated as a random event happening between the object of interest and the camera.
An optimization framework for 3D reconstruction of a transparent object viewed from different viewpoints on a turntable is shown in~\cite{wu2018full}.

A multi-view position-normal consistency was suggested in~\cite{qian20163d,qian2017stereo,qian2018simultaneous}.
The constraint imposes consistency on the normal field between the different views.
%
% Noteworthy, in our approach, we reconstruct the backward paths for the entire input and measure error on individual light paths for the entire image from which we estimate hypothesized normals for Snell's law.
%Note that, in our approach, we optimise the surface for the whole image and estimate normals from it, which avoids the point to point ambiguity and needs only a single image of the refracted surface and no additional priors.
%
In~\cite{qian2017stereo} this constraint was used to estimate a refractive surface in a stereo setting with a known background texture and depth.
In~\cite{qian2018simultaneous} this constraint was used for 3D shape reconstruction of a wavy water surface together with the refracted underwater scene from multiple views.

\noindent\textbf{Radiometric Approaches.}
Chari~\etal{}~\cite{chari2013theory} showed that radiometric clues in addition to the geometric distortion can help for refractive object reconstruction.
The method of Ihrke~\etal{}~\cite{ihrke2005reconstructing} used a level set approach to reconstruct a dynamic dyed liquid using a fluorescent substance.
In~\cite{asano2016shape} spectral dependency of water is used to reveal object geometry immersed in water from the absorption coefficient by applying Beer-Lambert law.

\noindent\textbf{Monocular Reconstruction.}
Pioneering work in monocular refractive surface reconstruction was introduced in~\cite{murase1992surface}. They presented an approach for non-rigid transparent shape reconstruction by statistical and geometrical clues from surface motion over a known pattern using tens of frames.
The work of Wetzstein~\etal{}~\cite{wetzstein2011refractive} uses specially designed patterns that can encode angular information of the scene and estimate surface geometry from a single image with the additional angular information provided by the pattern.

A recent learning-based method~\cite{stets2019single} uses a large set of synthetic images to train a model which can reconstruct complex refractive objects from a single image with a known distant background. Our method recovers the surface using the physics of refraction and does not rely on any training data.
Thapa~\etal{}~\cite{thapa2020dynamic} use a series of images \new{under orthographic projection} to learn the fluid dynamics of water and estimate a \new{height map} of a dynamic surface from images \new{with a known background}.

Similarly to our method,~\cite{shan2012refractive} they present an optimization approach in a monocular setting. However, they reconstruct height-field under an orthographic projection and require a known flat-patterned background. The parallel viewpoint direction in the orthographic projection eliminates the need to estimate the absolute distance of the surface, since moving the surface away from the camera does not influence refraction. Unlike~\cite{shan2012refractive}, our method estimates the absolute surface depth directly by optimizing reconstruction error on refracted light paths. % by applying Snell's law with a perspective camera and arbitrary background.
\new{Our energy optimizes over world coordinates instead of reprojection error in the image plane. This enables us to cope with the realistic setup of an arbitrary background shape, unlike the formulation of~\cite{shan2012refractive} which is limited solely to the flat background. Furthermore, optimization in world coordinates allows us to exploit the perspective projection of the refracted rays and therefore avoid the height-ambiguity present in~\cite{shan2012refractive,thapa2020dynamic}. }

%% file: tex/background.tex
\input{local_figures/schema}

Consider a camera whose reference frame is in the origin and pointing towards a positive $Z$ coordinate. The matrix of the intrinsic parameters $\calb{}$ of the camera is known. Furthermore, consider a known background point $\Xb{}$, see~\fig{fig:bg:schema} [left].
When the medium between point $\Xb{}$ and the camera is homogeneous, the path of the ray from the point to the observer (camera) forms a straight line, as is usually assumed in computer vision. Then we denote the image coordinate of $\Xb{}$ as $\xb{}$.

The situation changes when an object with a different refraction index is placed between the camera and the background.
Light passing through the interface between the two media is refracted and the image of the background is distorted.
We denote image coordinates of the refracted ray as $\xr{}$, which is a projection of the 3D point surface point $\Xr{}$ located on the refractive surface, see \fig{fig:bg:schema} [center].

% The location of $\Xr{}$ is given by Fermat's principle of the least time, \ie{}, $\Xr{}$ will be on the trajectory between the object $\Xb{}$ and its observation $\xr{}$ whose travel time is stationary.
%
In our setting, we assume only two homogeneous media, one medium where the observer is located with an index of refraction (IOR) $n_1$, the other where is the background with IOR $n_2$. 
With known IORs, we can apply Snell's law~\cite{kidger2001fundamental} which defines how refraction happens between two media.
Formally, let us denote in $\LOS{}$ the direction of the ray coming out of the camera and hitting the refractive surface at a point that has a normal $\n{}$.
Then, a ray exits the interface at direction $\s{}$, given by Snell's law~\cite{kidger2001fundamental}
\begin{equation}
\s{} = \snell{\LOS{}}{\n{}}{\mu} := -\mu \LOS{}+\n{}\cdot\left(\mu\alpha-\sqrt{1-\mu^{2}\left(1-\alpha^{2}\right)}\right)\;\;,
\label{eq:bg:snell}
\end{equation}
where $\mu=n_1/n_2$ is the ratio of the IORs and $\alpha=\LOS{}^T\n{}$.
By knowing the point $\xr{}$, we can calculate the line-of-sight (LOS) on which the point $\Xr{}$ must lie.
The LOS is an unprojection of a point $\xr{}$,
\begin{equation}
\begin{array}{ccc}
\LOSr{}\ \hat{=}\ \calb{}^+\xr{}\;\;.
\end{array}
\label{eq:bg:LOS}
\end{equation}

The critical yet straightforward fact is that the point $\Xr{}$ must lie on its corresponding observed LOS $\LOS{}$.
Given $\LOSr{}$ we know that for a distance $\de{}$ we can calculate a point $\Xr{}$
%
%With each observed point, to reconstruct the original 3D point we have one-dimensional subspace with just a single DOF defined as
\begin{equation}
\begin{array}{ccc}
\Xr{}(\de{})\ =\ \de{} \LOS{}\;\;,
\end{array}
\label{eq:bg:PC}
\end{equation}
\ie{}, $\Xr{}$ lies on the ray $\LOS{}$ at a distance of $\de{}$ from the camera origin where $\Xr{}=\left[x,y,z\right]^T$ is a surface point with a depth $z$.
Combining \eq{eq:bg:snell} and \eq{eq:bg:PC} yields the expression for the ray's light path that connects $\Xr{}$ and $\Xb{}$
\begin{equation}
\Xb{}=\Xr{} + t\s{}\;\;,
\label{eq:bg:lightpath}
\end{equation}
where $t$ is the distance from $\Xr{}$ to the background point $\Xb{}=\left[x^B,y^B,z^B\right]^T$, see \fig{fig:bg:schema} [center].

%% file: local_figures/schema.tex
\begin{figure*}[ht!]
    \centering
    \resizebox{1.0\linewidth}{!}{
    \begin{tabular}{ccc}
    \includegraphics[width=0.3\linewidth]{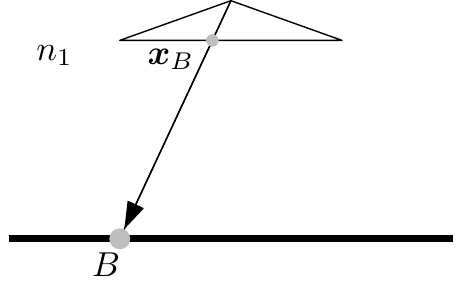} & 
    \includegraphics[width=0.3\linewidth]{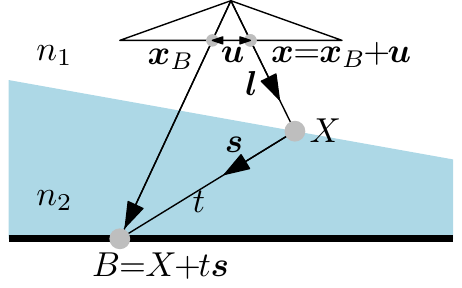} & 
    \includegraphics[width=0.3\linewidth]{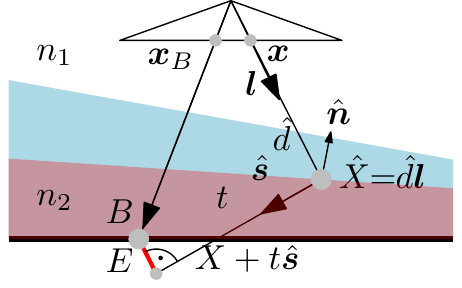}
    \end{tabular}}
    \caption{The geometry of monocular imaging through a surface.
    [Left]~Imaging without refraction. A background point at $\Xb{}$ is projected on an image plane at $\xb{}$ to an image plane with color $I(\xb{})$.
    [Center]~Imaging through a homogeneous refractive material between $\Xb{}$ and the camera.
    The light emanating from the background point $\Xb{}$ is refracted and projected to the coordinate $\xr{}$. In this case, the light path consists of two line segments that go from $\xr{}$ through $\Xr{}$ and meet with the background point $\Xb{}$.
    [Right]~Light path when the estimated refractive surface (red) is different from the correct one (blue).
    We measure the energy of a proposed solution $\len{}$ as the shortest distance between the estimated light path ($\hat{\Xr{}}+t\hat{\s{}}$, see \eq{eq:bg:LOS}) and $\Xb{}$ which should (ideally) be intersected and then $\len{}=0$.
    The direction of the refracted ray $\hat{\s{}}$ is calculated from the normal $\hat{\n{}}$ and the line-of-sight $\LOS{}$ and we know that the correct refracted surface point lies on $\LOS{}$.
    }
    \label{fig:bg:schema}
\end{figure*}

%% file: tex/optim.tex
\input{local_figures/energy}

The input to the method is $I_B$, an image of the background object acquired through a homogeneous medium (\ie{} air) including its depth $\Xb{}$ for each pixel, and $I_R$, an image of the background object acquired through the refractive surface from the same viewpoint. Then, dense matches are found between these two images, which is done by estimating the flow field $\OF{}$ such that $I_B(\xr{})=I_R(\xb{}+\OF{})$.

Once the dense matches $\OF{}$ matches are found, the optimization solves for the refractive surface by finding the $\dev{}$ along the known LOS $\LOS{}$ for each object point.
The optimization aims to find the values $\dev{}$ for all points such that their light paths will pass through their respective background points $\Xb{}$ (see \fig{fig:bg:schema} [center]).
%Our energy penalizes incorrect light path predictions by its minimal distance from the original background point.
%
%It is the smallest proximity between the backward reconstructed light path in \eq{eq:bg:lightpath} and the background point $\Xb{}$, see \fig{fig:bg:schema} [right].

In each iteration of the optimization, we perform the following steps:
1)~Calculating the surface point cloud (PC) using \eq{eq:bg:PC} with the current estimate of the distances $\dev{}$. 
2)~Calculating the normals $\n{}$ from the entire PC. 
3)~Calculating $\s{}$ using~\eq{eq:bg:snell} and reconstructing the estimated light path in~\eq{eq:bg:lightpath}.

The steps are performed in each update of $\dev{}$ until a stop criterion on the value of the energy is met.

\subsection{The Energy}
\vspace{-0.5em}
In this part, we describe the energy we optimize to obtain surface geometry.
We first show how we measure the error of the depth map when it does not form a correct light path.

In \sect{sect:bg} we said that $\Xb{}$ has to lie on the trajectory defined in \eq{eq:bg:lightpath} (see~\fig{fig:bg:schema}~[center]).
During the estimation of the surface, we reconstruct light paths using~\eq{eq:bg:lightpath}, which should ideally intersect with $\Xb{}$.
For an incorrect $\de{}$, the path does not intersect with its $\Xb{}$, and we penalize the estimate by its smallest distance of the light path to its $\Xb{}$.
Note that we have to reconstruct light paths for the entire image since normals are needed to calculate $\s{}$ in~\eq{eq:bg:snell}.

The situation is schematically shown in~\fig{fig:bg:schema}~[right], where the light path does not intersect with the background point $\Xb{}$.
%
% In that case, the energy we minimize is calculated from a point on the light path in \eq{eq:bg:LOS} along $\s{}$ with the smallest proximity from its corresponding background point $\Xb{}$.
In this case, the energy of a point is calculated as the shortest distance between a light path in \eq{eq:bg:LOS} and its background point, $\Xb{}$. This point-line distance has a following closed form,
\begin{equation}
\len{}(\X{},\s{},\Xb{}) := \left \|\Xb{} - \left[\Xr{} + [\s{}\cdot(\Xb{} - \Xr{})] \s{}\right]\right\|_2
\label{eq:optim:dist}
\end{equation}
\noindent where $\cdot$ is a dot product and $\Xr{}$ is given by \eq{eq:bg:PC}.
The total energy we optimize for all points jointly is
\begin{equation}
\dev{}^* = \arg \min_{\dev{}} {\sum_{i\in\Omega}\len{}(\Xr{i},\s{}\left(\LOS{i},\n{i},\mu\right),\Xb{i})} + \max\{0,z_{i} - z_{i}^{B}\}\;\;,
\label{eq:optim:energy}
\end{equation}
where individual normals $\n{i}$ are calculated from all points in PC $\Omega{}$ of all surface points $\Xr{i}(\dev{})$ in~\eq{eq:bg:PC}. The second term prevents estimated depth $z_i$ to fall behind the background depth $z_{i}^B$.
Note that the normals and PC are calculated in the iteration of the optimization step, but only $\dev{}$ is updated.

\fig{fig:optim:energy} shows a simple example of our energy \eq{eq:optim:dist} with three different combinations of point distances of two adjacent points.
The energies are calculated on a grid of distances. Each image shows that even when only one of the points is correct, the energy is negatively affected as the points are spatially linked by normals. This link imposed by normals affects neighbouring points, and the lower energy is always at the location which matches the true value for both points. \new{The spatial linking through normals introduces the only regularization in our energy. It imposes physical correctness on the entire depth map because it is linked to other surface points through normals which are input for Snell's law. }
\new{Moreover, the figure reveals that there is a significant drop in the gradient magnitude when a solution is either close to a correct solution or its scalar multiple of depth, which corresponds to the ridge in the energy surface shown in \fig{fig:optim:energy}.}

\noindent\textbf{Initialization.}
\label{sect:init}
The energy is non-convex and relies on a good initial guess, like coarse information about the surface distance.
For this purpose, we developed an initialization scheme where we optimize a location of a plane perpendicular to the camera's principal axis. The plane location from the observer is defined by a single variable, a shift, which we can easily and quickly optimize with the energy, \new{\ie{}, we optimize for $c$ after plugging $\dev{flat} = c$ as the depth to the energy in~\eq{eq:optim:energy}}.
With the initial estimate, the location of the plane is usually roughly located around the mean position of the ground truth surface and serves as a good initial guess.

%% file: local_figures/energy.tex
\newcommand{\makeenergyimagecell}[2]{\begin{overpic}[width=0.3\linewidth]{local_images/energy/#1} \put (5,65) {\small #2} \end{overpic}}

\begin{figure*}
    \centering
    \resizebox{1.0\textwidth}{!}{
    \begin{tabular}{ccc}
        \makeenergyimagecell{15_25}{$\de{1} = 1.5, \de{2} = 2.5$} & 
        \makeenergyimagecell{25_15}{$\de{1} = 2.5, \de{2} = 1.5$} & 
        \makeenergyimagecell{25_25}{$\de{1} = 2.5, \de{2} = 2.5$}
    \end{tabular}}\vspace{0.1cm}
    \caption{
    A minimal synthetic example of the proposed energy $\len{}$ for a two-point surface $\de{1},\de{2}$.
    The background plane is placed at $\Xb{}^z=3$ and the shown surfaces are energies of different combinations of $\de{1},\de{2}$ with their ground truth location shown above.
    The values are the logarithm of $\len{}$ for better visibility.
    All energies show that the ground truth solution yields a global minimum. However, notice that the magnitude of the gradient gets very small around the location of the correct solution. Similarly, there is a ridge along a scalar multiple of the depth. 
    The peak in the energy slightly deviated from the ground truth, because the normals are calculated from the numerical approximation of surface derivatives.
    }
    \label{fig:optim:energy}
\end{figure*}

%% file: tex/experiments.tex
To showcase our method, we compare it quantitatively to benchmarks with available ground truth ~\cite{qian2018simultaneous,thapa2020dynamic} and make a series of real-world experiments to estimate a surface of a liquid with a standard off-the-shelf camera.
\new{We have chosen a CPU Matlab implementation of L-BFGS~\cite{zhu1997algorithm} which does not require the calculation of the memory-prohibitive Jacobian or Hessian as the number of the variables equals the number of surface points.}
\new{Run time of one energy evaluation is on average $2.06$ ms for 64 $\times$ 64 px images, $7.26$~ms for $128\times128$ px and $25.32$ ms for $256 \times 256$ px images.}
For dense matches between background and refracted surface $\OF{}$ we used the optical flow~\cite{brox2010large}.
We optimize the energy for the entire input surface simultaneously, but alternatively, the optimization step can be performed patch-wisely.
The patch-wise computation can however hurt the quality in non-overlapping regions.

\noindent\textbf{Synthetic Experiments.} 
We compare our method to two methods, a multiview method~\cite{qian2018simultaneous} and a monocular method~\cite{thapa2020dynamic}.

In the first experiment, we use the same benchmark functions presented in~\cite{qian2018simultaneous} with a perspective projection.
The benchmarks are denoted \benchmarkname{wave1} and \benchmarkname{wave2}.
The surface~\benchmarkname{wave1} is defined as \mbox{$\zone(x,y,t) = 2+0.1\cos[{\pi{} (t + 50) \sqrt{(x-1)^2 + (y - 0.5)^2}}/{80}]$}.
It can be intuitively seen as a diagonally rolling wave.
The second benchmark \benchmarkname{wave2}, $\ztwo(x,y,t) = 2-0.1\cos[{\pi{}(t+60)\sqrt{ (x + 0.05)^2 + (y+0.05)^2}}/{75}]$ can be seen as a wave growing in the center.
Since surfaces are smooth, a resolution of $64\times64$ px is sufficient to represent the surface.
Scenes are rendered with Blender~\cite{blender} with Cycles Engine. The rendering scripts and input images will be made available with the paper.
Both functions are evaluated on $100$ surfaces in $t\in(0,99)$.
We tested both benchmark functions with two background shapes to show that background geometry can be arbitrary. The first, a plane at \mbox{$\zflat{}=2.5$} and the second, a more complex shape, \mbox{$\zfunc{}(x,y) = 2.5 + 0.05 [\sin(2\pi{} x) + \cos(2\pi{}y)]$}.

The results of the first experiment are shown upper part of~\tab{tab:experiments:MSE} and~\fig{fig:experiment_waves}. 
We show results for three initialization schemes, the independent initialization is done for each frame $t$ independently and sequential initialization uses the previous estimate as the initial value, which is similar to the technique used in~\cite{qian2018simultaneous}. 
Lastly, to demonstrate that our method works well when an initial surface location is provided, we initialized the method with the surface location at $\dev{0} = 2$, which roughly corresponds to a mean location of the sought surface.
Error is reported in both Root Mean Square Error ($\RMSE{}$) of the reconstructed depth map and Mean Angular Error ($\MAE{}$) of the estimated normals. \new{We used the same calibration as~\cite{qian2018simultaneous} therefore the $\RMSE{}$ is measured in the same (arbitrary) \unit{}s}.
\input{local_figures/mse}
\input{local_figures/experiment_waves}
\input{local_figures/experiment_thapa}
%
%Quantitatively, on the synthetic data $\benchmarkname{wave1}$ and $\benchmarkname{wave2}$ our method performs worse than the existing multiview approach~\cite{qian2018simultaneous}.
%
%Our method, however, unlike~\cite{qian2018simultaneous}, completely relies on a monocular setting with just predefined depth and texture of the background object.
%
The errors vary with time as the surfaces become more complex, but the shape of the background surface has little influence on the results. 
We see that our method is sensitive to initialization. There is a noticeable difference in $\RMSE$ when a good initial value is provided ($\dev{0} = 2$, roughly the location of the correct surface).
Results differ particularly on \benchmarkname{wave2}, where $\RMSE{}$ is approximately six times better when a good initialization is used.
A less noticeable gap in results is apparent on \benchmarkname{wave1} where $\MAE{}$ is even better when sequential initialization is used than in the case of the good initial guess ($\dev{0}=2$) but still worse in $\RMSE{}$.
Note that we rely on a monocular setting with just a predefined background.
In comparison with the nine-camera setting of~\cite{qian2018simultaneous}, our method performs worse both in $\RMSE{}$ and $\MAE{}$.
Since the benchmark surfaces are largely non-flat, our method sometimes estimated the mean location of the surface slightly off but recovered the normals comparatively accurately to the four camera settings in~\cite{qian2018simultaneous} on \benchmarkname{wave1} benchmark.
However, in the reduced setting with four cameras~\cite{qian2018simultaneous}, our method is on par in $\MAE{}$, but is still worse in $\RMSE{}$.

\new{The second experiment compares our method to~\cite{thapa2020dynamic} in a monocular setting with orthographic projection.
%
%The benchmark data~\cite{thapa2020dynamic} do not provide optical flow and some images contain very little texture to estimate it, therefore we generated surfaces with their code. The scenarios are not identical to their benchmark but we used the parameters they provided and reproduced very similar surfaces. 
We used the code provided by~\cite{thapa2020dynamic} to generate the surfaces and their optical flow.
We rendered ten frames of the three available scenarios \benchmarkname{ocean}, \benchmarkname{ripple} and \benchmarkname{tian} with $128\times128$ px under orthographic projection.
Note that there is a height-ambiguity imposed by the orthographic projection, as
a shift of the entire surface does not change the refracted ray $\s{}$ since the angle subtended between $\LOS{}$ and $\n{}$ does not change with by moving the surface in $Z$ coordinate, therefore the estimated surfaces are normalized to have zero mean before their $\RMSE{}$ is calculated. For the same reason, all frames were initialized identically with an arbitrary flat surface at $\dev{0}=1$.%

Our method quantitatively performs better in $\RMSE{}$ than~\cite{thapa2020dynamic} in a single frame setting.  
For the full multi-frame FSRN-RNN~\cite{thapa2020dynamic} the $\RMSE{}$ error is half compared to ours but this method takes into account fluid dynamics between subsequent frames, unlike our method which works independently on each frame. 
It is important to stress that in~\cite{thapa2020dynamic} a flat background at $z = 0$ is required, while in our case the background depth can be arbitrary. 
Furthermore, \cite{thapa2020dynamic,shan2012refractive} are limited only to orthographic projection and thus can only estimate height-field, while our approach is more general and can estimate a 3D surface under the perspective setting.  }
%Our method however depends on optical flow which must be provided beforehand.

\new{In both experimental setups, our method is efficient in finding the correct surface shape as is evident in the low $\MAE{}$. Most of the $\RMSE{}$ error stems from errors in the shift and scale of the surface, as discussed in~\sect{sect:optim}}.

\input{local_figures/experiment_real_img}
\input{local_figures/experiment_real}

\noindent\textbf{Real-World Experiments.}
We further evaluated our approach on real-world scenes using a water tank and a flat surface with a printed complex texture as the background  (shown in~\fig{fig:experiment_water_img} [left]). We used a Nikon D810 camera with a Nikon AF-S Nikkor 105 mm lens.
Camera calibration was performed with the MATLAB Camera Calibration Toolbox. We also used the results of the depth of the intrinsic calibration marker to measure the depth of the flat background surface. The optical flow was estimated with~\cite{brox2010large}.
We fixed the camera to a steel custom-made structure approximately 630 mm away from the background (an averaged distance of the flat calibration pattern) where the background is perpendicularly aligned to the principal axis of the camera.
The textures contain many fine details, which are particularly good for finding the dense matches between the background image and the images distorted by refraction. We show the background and one refracted input image in~\fig{fig:experiment_water_img}~[left] and the estimated surfaces in~\fig{fig:experiment_water}.
%

%Results show that surfaces are mostly flat, therefore the aforementioned plane fitting initialisation scheme works well.
%
We perturbed the surface by hand to make different types of waves with varying smoothness.
Frames 1 and 2 in~\fig{fig:experiment_water} show reconstructions of smooth surfaces with rather larger waves.
Frame 4 showcases a situation with a slightly larger depth range where we shook the surface.
Lastly, frames 5 and 6, in contrast, show waves with rather sharper edges of smaller depths.
Since the real-world surface is mostly flat, the initial estimate is often accurate enough and produces convincing depths.

%% file: local_figures/mse.tex
\begin{table}[t]
\resizebox{1.0\linewidth}{!}{
    \begin{tabular}{l|cc|cc|cc|cc|cc}
         \multicolumn{11}{c}{Perspective}\\
         & \multicolumn{2}{c|}{Ours - indep. init} & 
         \multicolumn{2}{c|}{Ours - seq. init} & 
         \multicolumn{2}{c|}{Ours - $\dev{0} = 2$} &
         \multicolumn{2}{c|}{$9$ Cameras~\cite{qian2018simultaneous}} & 
         \multicolumn{2}{c}{$4$ Cameras~\cite{qian2018simultaneous}}\\
         \hline
         & \RMSE{} [\unit{}] & \MAE{} [$^\circ$] & 
         \RMSE{} [\unit{}] & \MAE{} [$^\circ$] & 
         \RMSE{} [\unit{}] & \MAE{} [$^\circ$] & 
         \RMSE{} [\unit{}] & \MAE{} [$^\circ$] &
         \RMSE{} [\unit{}] & \MAE{} [$^\circ$]\\
         \hline
         \hline
         \benchmarkname{wave1}-$\zflat$ & 
         \new{0.15} & $5.89^{\circ}$ & 
         \new{0.11} & $4.29^{\circ}$ & 
         \new{0.05} & $5.06^{\circ}$ &
         \multirow{2}{*}{0.006} & \multirow{2}{*}{$0.76^{\circ}$} & 
         \multirow{2}{*}{$0.014$} & \multirow{2}{*}{$4^{\circ}$}\\
         \benchmarkname{wave1}-$\zfunc$ & 
         \new{0.14} & $ 6.06^{\circ}$ & 
         \new{0.10} & $4.79^{\circ}$ & 
         \new{0.05} & $4.80^{\circ}$ & & &\\
         \benchmarkname{wave2}-$\zflat$ & 
         \new{0.23} & $11.16^{\circ}$ & 
         \new{0.45} & $ 13.28^{\circ}$ & 
         \new{0.04} & $5.89^{\circ}$ & 
         \multirow{2}{*}{0.002} & \multirow{2}{*}{$0.37^{\circ}$} & 
         \multirow{2}{*}{-} & \multirow{2}{*}{-}\\
         \benchmarkname{wave2}-$\zfunc$ & 
         \new{0.23} & $ 10.85^{\circ}$ & 
         \new{0.43} & $ 13.18^{\circ}$ & 
         \new{0.04} & $5.92^{\circ}$ & & &\\
         \hline
         \hline
         
         \multicolumn{11}{c}{\new{Ortographic}}\\
         & \multicolumn{6}{c|}{\new{Ours - indep. init}} &
         \multicolumn{2}{c|}{\new{Full~\cite{thapa2020dynamic}}} & 
         \multicolumn{2}{c}{\new{Single-input~\cite{thapa2020dynamic}}}\\
         \hline
         & \multicolumn{3}{c}{\new{\RMSE{}}} & \multicolumn{3}{c|}{\new{\MAE{} [$^\circ$]}} &
         \new{\RMSE{}} & \new{\MAE{} [$^\circ$]} &
         \new{\RMSE{}} & \new{\MAE{} [$^\circ$]}\\
         \hline
         \hline
         \new{\benchmarkname{ocean}} & \multicolumn{3}{c}{\new{0.230}} & \multicolumn{3}{c|}{\new{$1.301^{\circ}$}} & \multirow{3}{*}{\new{0.126}} & \multirow{3}{*}{\new{-}} & \multirow{3}{*}{\new{0.262}} & \multirow{3}{*}{\new{-}} \\
         \new{\benchmarkname{ripple}} & \multicolumn{3}{c}{\new{0.217}} & \multicolumn{3}{c|}{\new{$0.649^{\circ}$}} & & & & \\
         \new{\benchmarkname{tian}} & \multicolumn{3}{c}{\new{0.248}} & \multicolumn{3}{c|}{\new{$1.364^{\circ}$}} & & & & \\
         \hline
    \end{tabular}} \vspace{0.01cm}
    \caption{
    Quantitative comparison of reconstructed surfaces with a multi-view~\cite{qian2018simultaneous} \new{and a monocular method~\cite{thapa2020dynamic} }. 
    The upper part of the table shows results on two benchmark functions \benchmarkname{wave1} and \benchmarkname{wave2} in perspective setting.
    To demonstrate that our method can handle different background depths, we conducted experiments with two different background depths defined by~$\zflat{}$ and~$\zfunc{}$.
    The table shows three different initialization schemes, the first column shows results when each frame is initialized separately, the second the sequential initialization when only a first frame is initialized and consecutive frames use the previous result and the third is where all frames use the same initial $\dev{0}=2$.
    \new{The lower part shows evaluation of three different surfaces (\benchmarkname{ocean}, \benchmarkname{ripple} and \benchmarkname{tian}) compared to~\cite{thapa2020dynamic}  (monocular setting under orthographic projection).
    Each of these surfaces consists of a ten-frame sequence generated with the code provided by~\cite{thapa2020dynamic}.}
    }\vspace{-0.1cm}
    \label{tab:experiments:MSE}
\end{table}

%% file: local_figures/experiment_waves.tex
\begin{figure}[t]
    \centering
    \resizebox{1.0\linewidth}{!}{
    \begin{tabular}{ccccccc}
        \rotatebox{90}{\hspace{1em}estimated} &
         \makewaveimagecell{wave1/complex/est/0025.pdf}{\benchmarkname{wave1}, $t=25$} & 
         \makewaveimagecell{wave1/complex/est/0050.pdf}{\benchmarkname{wave1}, $t=50$} & 
         \makewaveimagecell{wave1/complex/est/0075.pdf}{\benchmarkname{wave1}, $t=75$} & 
         \makewaveimagecell{wave2/complex/est/0025.pdf}{\benchmarkname{wave2}, $t=25$} & 
         \makewaveimagecell{wave2/complex/est/0050.pdf}{\benchmarkname{wave2}, $t=50$} & 
         \makewaveimagecell{wave2/complex/est/0075.pdf}{\benchmarkname{wave2}, $t=75$}\\
         \rotatebox{90}{\hspace{1em}ground truth} &
         \makewaveimagecell{wave1/gt/gt_0025.pdf}{} & 
         \makewaveimagecell{wave1/gt/gt_0050.pdf}{} & 
         \makewaveimagecell{wave1/gt/gt_0075.pdf}{} &
         \makewaveimagecell{wave2/gt/gt_0025.pdf}{} & 
         \makewaveimagecell{wave2/gt/gt_0050.pdf}{} & 
         \makewaveimagecell{wave2/gt/gt_0075.pdf}{}\\
         \rotatebox{90}{\hspace{2em}RMSE{}} &
         \makewaveimagecell{wave1/complex/err/0025.pdf}{} & 
         \makewaveimagecell{wave1/complex/err/0050.pdf}{} & 
         \makewaveimagecell{wave1/complex/err/0075.pdf}{} & 
         \makewaveimagecell{wave2/complex/err/0025.pdf}{} & 
         \makewaveimagecell{wave2/complex/err/0050.pdf}{} & 
         \makewaveimagecell{wave2/complex/err/0075.pdf}{}\\
    \end{tabular}}\vspace{0.1cm}
    \caption{Results of the \benchmarkname{wave1} and \benchmarkname{wave2} with $\zfunc{}$ background at times $t\in\{25,50,75\}$. Initialization was done for each frame independently.
    Estimated [top] and ground truth surfaces [middle] with colour-coded~\cite{baker2011database} optical flow~\cite{brox2010large}. \new{The bottom row shows a colour coded error of the estimated depth map.}}\vspace{-0.1cm}
    \label{fig:experiment_waves}
\end{figure}

%% file: local_figures/experiment_thapa.tex
\newcommand{\makethapaimagecell}[2]{\begin{overpic}[width=0.3\linewidth]{local_images/experiments/#1} \put (0,75) {{\huge\benchmarkname{#2}}} \end{overpic}}

\begin{figure}[t]
    \vspace{1em}
    \centering
    \resizebox{1.0\linewidth}{!}{
    \begin{tabular}{ccccccc}
         % \rotatebox{90}{\hspace{2em}estimated} &          \makethapaimagecell{thapa/ocean/est/Seq_1_5_1}{ocean $t=1$} & \makethapaimagecell{thapa/ocean/est/Seq_1_5_10}{ocean $t=10$} & \makethapaimagecell{thapa/ripple/est/Seq_1_5_1}{ripple $t=1$} &          \makethapaimagecell{thapa/ripple/est/Seq_1_5_10}{ripple $t=10$} & \makethapaimagecell{thapa/tian/est/Seq_1_5_1}{tian $t=1$} &          \makethapaimagecell{thapa/tian/est/Seq_1_5_10}{tian $t=10$}\\
         \rotatebox{90}{\hspace{2em}\Large{estimated}} &          
         \makethapaimagecell{thapa/ocean/est/Seq_1_5_1}{ocean $t=1$} &          \makethapaimagecell{thapa/ocean/est/Seq_1_5_10}{ocean $t=10$} &          \makethapaimagecell{thapa/ripple/est/Seq_1_5_1}{ripple $t=1$} &          \makethapaimagecell{thapa/ripple/est/Seq_1_5_10}{ripple $t=10$} &         \makethapaimagecell{thapa/tian/est/Seq_1_5_1}{tian $t=1$} &         \makethapaimagecell{thapa/tian/est/Seq_1_5_10}{tian $t=10$}\\
         \rotatebox{90}{\hspace{1em}\Large{ground truth}} &
         \makethapaimagecell{thapa/ocean/gt/gt_Seq_1_5_1}{} &
         \makethapaimagecell{thapa/ocean/gt/gt_Seq_1_5_10}{} & 
         \makethapaimagecell{thapa/ripple/gt/gt_Seq_1_5_1}{} & 
         \makethapaimagecell{thapa/ripple/gt/gt_Seq_1_5_10}{} & 
         \makethapaimagecell{thapa/tian/gt/gt_Seq_1_5_1}{} &
         \makethapaimagecell{thapa/tian/gt/gt_Seq_1_5_10}{}\\
         \rotatebox{90}{\hspace{2em}\Large{\RMSE{}}} &          
         \makethapaimagecell{thapa/ocean/err/Seq_1_5_1}{} &          \makethapaimagecell{thapa/ocean/err/Seq_1_5_10}{} &          \makethapaimagecell{thapa/ripple/err/Seq_1_5_1}{} &          \makethapaimagecell{thapa/ripple/err/Seq_1_5_10}{} &         \makethapaimagecell{thapa/tian/err/Seq_1_5_1}{} &         \makethapaimagecell{thapa/tian/err/Seq_1_5_10}{}
    \end{tabular}}\vspace{0.15cm}
    \caption{
   \new{Estimated surfaces of \benchmarkname{ocean}, \benchmarkname{ripple}, \benchmarkname{tian} of surfaces generated from~\cite{thapa2020dynamic} under an ortogtraphic projection.
    Estimated [top] and ground truth surfaces [middle] with colour-coded~\cite{baker2011database} given optical flow~\cite{brox2010large}. 
    The bottom row shows a colour coded error of the estimated depth map.
    Due to the height-ambiguity imposed by the problem, the comparison is performed with surfaces normalized to have a zero mean.}    }
    \label{fig:experiment_thapa}
\end{figure}

%% file: local_figures/experiment_real_img.tex
\begin{figure}[t]
    \centering
    \resizebox{1.0\linewidth}{!}{
    \begin{tabular}{ccc}
        % Background image & Input Frame 8\\
        \includegraphics[width=0.3\linewidth,angle=180]{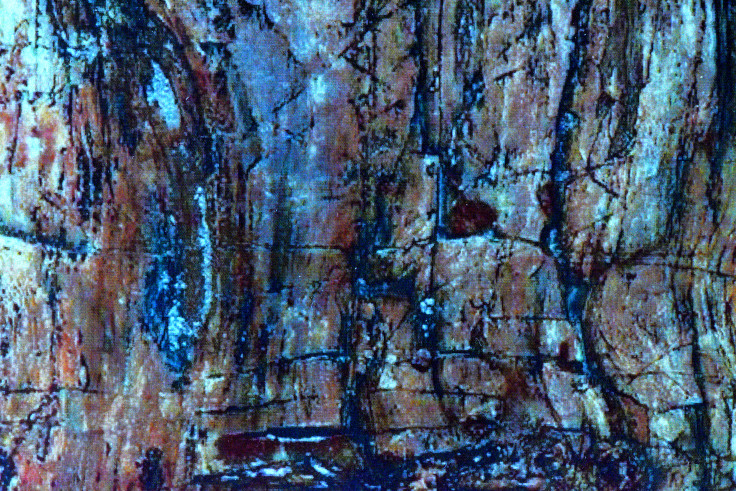} & \includegraphics[width=0.3\linewidth,angle=180]{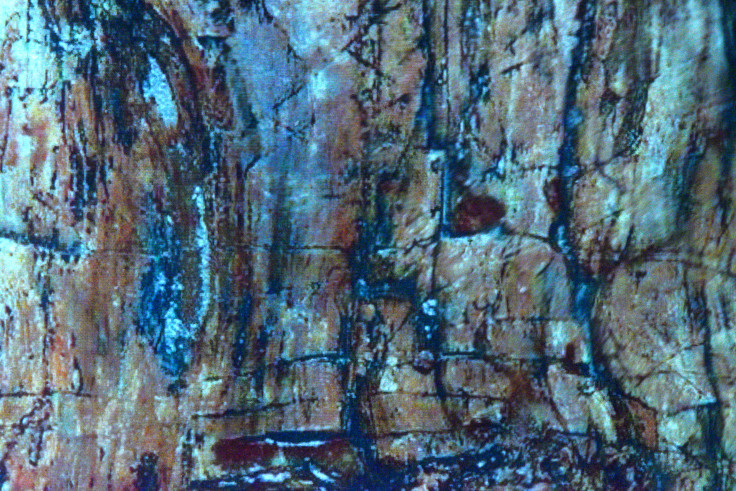} & \scalebox{-1}[1]{\includegraphics[width=0.3\linewidth,angle=180]{local_images/experiments/bark_full/of/RGT_5143.png}}
    \end{tabular}}\vspace{0.1cm}
    \caption{[Left]~The input background texture we used in our real-world experiment. [Center]~One of the input images of a refracted surface (denoted as Frame 6 in \fig{fig:experiment_water}). \new{[Right]~Optical flow between these frames.}
    % 
    %Supplementary material contains animation of all input images. 
    }
    \label{fig:experiment_water_img}
\end{figure}

%% file: local_figures/experiment_real.tex
% this shows the optical flow in the left upper corner
% \newcommand{\makerealimagecell}[2]{\begin{overpic}[width=0.55\linewidth]{local_images/experiments/bark_full/#1} \put (0,80) {\scalebox{1}[-1]{\includegraphics[width=0.20\linewidth]{local_images/experiments/bark_full/of/#1}}} \put(70,70) {Frame #2}\end{overpic}}

% this does not show optical flow
\newcommand{\makerealimagecell}[2]{\begin{overpic}[width=0.4\linewidth]{local_images/experiments/bark_full/result/#1} \put(70,70) \Huge{Frame #2}\end{overpic}}

\newcommand{\rowskip}{\vspace{0em}}

\begin{figure}
    \centering
    \resizebox{1.0\linewidth}{!}{
    \begin{tabular}{ccc}
         \makerealimagecell{RGT_4920}{1} &  \makerealimagecell{RGT_4953}{2} & \makerealimagecell{RGT_4983}{3}\\
         \makerealimagecell{RGT_5006}{4} & \makerealimagecell{RGT_5123}{5} & \makerealimagecell{RGT_5143}{6}
    \end{tabular}
    }
    % \resizebox{1.0\textwidth}{!}{\begin{tabular}{cccc}
    % \makerealimagecell{RGT_4920}{1} &  \makerealimagecell{RGT_4953}{2}\\
    % \rowskip & \rowskip & \rowskip & \rowskip\\
    % \makerealimagecell{RGT_4983}{3} & \makerealimagecell{RGT_5006}{4}\\
    % \rowskip & \rowskip & \rowskip & \rowskip\\
    % \makerealimagecell{RGT_5123}{5} & \makerealimagecell{RGT_5143}{6}
    % \end{tabular}}
    \caption{Estimated surfaces of our real-world water experiment. \fig{fig:experiment_water_img} shows input images.
    Surfaces are shown with their corresponding colour-coded~\cite{baker2011database} optical flow~\cite{brox2010large}.
    The experiment was made with a water tank with a fixed camera acquiring the scene from above.
    Apart from making artificial waves on the surface, the amount of liquid was also changed to vary the overall depth.
    Because the magnitude of depth is significantly smaller than surface width and height, the shown surface depths are re-scaled 20 times for better visibility.}
    \label{fig:experiment_water}
\end{figure}

%% file: tex/conclusion.tex
A refractive object can hinder many conventional computer vision methods. Nevertheless, refraction can teach us about the physics of the scene and pose interesting challenges. 
Here we looked at the task of reconstructing a refractive surface over a known background using just a single view. This setup was never tackled before in the general setup of a perspective camera and an arbitrary background shape. 

Unlike previous papers that used multiple views, frames, or additional constraints, we showed that a simple energy function based on Snell's law can estimate a surface  under a monocular setup \new{when background depth is given}. Working directly at world coordinates, we express our energy as a distance between the original background point and the expected ray's trajectory. 
The energy looks for the surface that best explains the refracted image altogether. By optimizing for the entire surface, implicit smoothness is imposed, and the surface shape can be reconstructed.

We demonstrate our method on numerous synthetic and real-world scenes with varying complexity. 
We achieve errors that are on par in $\MAE{}$ with a previous method that used four views \new{and similarly accurate in $\RMSE{}$ with a deep learning single-frame monocular setup under orthographic projection.} 
Since our method is indirectly optimizing the depth through normal fields, it is highly sensitive to the initial depth and performs better in reconstructing surface normals than reconstructing absolute depth.
For initialization, we developed a simple initialization scheme, which works robustly on mostly flat surfaces. 
Our method is based on calculating the optical flow between the refracted image to the original one. Therefore, it is limited to scenarios where the optical flow can be reliably calculated, i.e., enough texture and moderate distortions from the refraction.
% In the future, we plan to adapt our energy to a multi-frame setting and jointly estimate the refractive and the background surface. 

%% file: tex/ack.tex
This work was supported by the grant ``Internationalisierung Exzellenzstrategie'' from University of Konstanz, ``SFB Transregio 161 - Quantitative Methods for Visual Computing (project B5)'', Leona M. and Harry B. Helmsley Charitable Trust, The Maurice Hatter Foundation, Israel Science Foundation grant $\#680/18$ and the Technion Ollendorff Minerva Center for Vision and Image Sciences. 